\documentclass{article}

\usepackage[normalem]{ulem}
\usepackage{graphicx} 
\usepackage{pifont}
\usepackage{balance}
\usepackage{multirow}
\usepackage{enumitem}
\usepackage{lipsum}
\usepackage{xspace}
\usepackage[]{hyperref}
\usepackage{amsmath}
\usepackage{algorithm}
\usepackage{algpseudocode}
\usepackage{tabularx}
\usepackage{booktabs}

\usepackage{multirow}
\usepackage{listings}
\usepackage{xcolor}

\usepackage[numbers,sort&compress]{natbib}
\setcitestyle{numbers,square, citecolor=green}

\hypersetup{
    colorlinks=true,    
    linkcolor=red,      
    citecolor=green,    
    filecolor=magenta,  
    urlcolor=magenta       
}

\renewcommand{\cite}{\citep}

\usepackage[utf8]{inputenc}
\usepackage[T1]{fontenc}
\usepackage{lmodern}

\usepackage[margin=1in]{geometry}
\usepackage{setspace}
\newcommand{\feh}[1]{\textcolor{black}{#1}}

\newcommand{\CL}[1]{\textcolor{black}{#1}}

\newcommand{\sys}{AgentRR\xspace} 

\begin{document}

\title{Get Experience from Practice: LLM Agents with Record \& Replay}


\author{Erhu Feng, Wenbo Zhou, Zibin Liu, Le Chen, Yunpeng Dong, Cheng Zhang, \\
        Yisheng Zhao, Dong Du, Zhichao Hua, Yubin Xia$^*$, Haibo Chen \\
	\emph{Institute of Parallel and Distributed Systems (IPADS)}, \\
	\emph{Shanghai Jiao Tong University}
} 

\maketitle 
\begin{abstract}

AI agents, empowered by Large Language Models (LLMs) and communication protocols such as MCP and A2A, have rapidly evolved from simple chatbots to autonomous entities capable of executing complex, multi-step tasks, demonstrating great potential.
However, the LLMs' inherent uncertainty and heavy computational resource requirements pose four significant challenges to the development of safe and efficient agents: reliability, privacy, cost and performance.
Existing approaches, like model alignment, workflow constraints and on-device model deployment, can partially alleviate some issues but often with limitations, failing to fundamentally resolve these challenges.

This paper proposes a new paradigm called \textbf{\sys} (Agent \emph{Record \& Replay}), which introduces the classical \emph{record-and-replay} mechanism into AI agent frameworks.
The core idea is to:
\textcircled{1} \textbf{record} an agent’s interaction trace with its environment and internal decision process during task execution,
\textcircled{2} \textbf{summarize} this trace into a structured ``\emph{experience}’’ encapsulating the workflow and constraints, and 
\textcircled{3} \textbf{replay} these experiences in subsequent similar tasks to guide the agent’s behavior.
We detail a multi-level \emph{experience} abstraction method and a \emph{check function} mechanism in \sys: the former balances experience specificity and generality, while the latter serves as a trust anchor to ensure completeness and safety during replay.
In addition, we explore multiple application modes of \sys, including user-recorded task demonstration, large-small model collaboration and privacy-aware agent execution, and envision an \emph{experience repository} for sharing and reusing knowledge to further reduce deployment cost.

While \sys exhibits significant potential to address current agent problems, 
it also introduces new challenges regarding experience completeness, replay robustness, generalization, and scope of applicability.
Therefore, this paper thoroughly analyzes the feasibility and advantages of \sys's mechanisms, 
and demonstrates its practical applications across common agent scenarios.
Furthermore, we establish theoretical foundations and propose novel solutions for future research in agent systems.

\end{abstract}

\begin{sloppypar}
\section{Introduction}

AI agents have seen rapid advancement\cite{liu2025}, evolving from basic conversational bots to intelligent entities capable of understanding complex instructions, autonomously planning\cite{chen2025, lin2025}, and executing multi-step tasks\cite{chhikara2025, goldie2025}.
Modern agents\cite{AnthropicComputerUse2025, reed2022, feng2025, zhou2023} demonstrate strong task-handling abilities across diverse domains, e.g., a PC-based agent\cite{UFO22025} autonomously conducting research and generating web pages, or a mobile assistant\cite{wang2025mobileagente} automatically finding coupon codes while shopping.
They can invoke various software tools via agent-to-tools communication protocols like MCP\cite{AnthropicMCP2025, hou2025mcp} and can coordinate with other agents through agent-to-agent protocols like A2A\cite{habler2025}. 
For example, the Manus AI system\cite{shen2025} can execute complex tasks like compiling to-do lists or searching real estate listings, and MobileAgent can perform intricate operations on smartphones. 
This growing autonomy endows agents with unprecedented capabilities, allowing them to independently plan and carry out complex actions.

However, such autonomy also means an agent's decision-making process can become opaque and unpredictable.
If an agent deviates from expected behavior, it may lead to serious consequences, heightening the need for reliability and controllability.
In other words, improving an agent's intelligence must go hand-in-hand with mechanisms to constrain and guide its behavior to ensure it stays on track.
Despite significant progress, widespread adoption of AI agents is hindered by four core challenges:
\begin{enumerate}

\item \textbf{Challenge-1: Reliability.}
LLM-based AI agents often exhibit the \emph{hallucination} phenomenon,
which is inherent in the probabilistic nature of LLMs\cite{Huang2025, banerjee2024, CometLLMHallucination2025, xu2025}.
An agent may function well most of the time, but without guaranteed correctness it can make unstable or incorrect decisions.
This unpredictability erodes trust: even powerful LLMs cannot ensure absolutely reliable agent behavior.

\item \textbf{Challenge-2: Privacy.}
Many agents, especially those backed by cloud-hosted LLMs, require users to send large amounts of personal or sensitive data\cite{wang2025, wu2025} (e.g. screenshots, documents, interaction logs) to remote servers for processing.
This raises serious privacy concerns: any cloud-based analysis of user data risks potential data leakage or misuse\cite{kim2025llmsonlineemergingthreat, chen2025obviousinvisiblethreatllmpowered, pinachodavidson2025proposalevaluatingoperationalrisk}.

\item \textbf{Challenge-3: Operational Cost.}
Sophisticated tasks typically demand multiple turns of interaction between the agent and the LLM\cite{li2025survey, goldie2025}.
Multi-modal tasks (e.g. analyzing GUI screenshots) tend to be even more expensive\cite{OpenAIPlatformPricing2025, GoogleGeminiPricing2025}, since vision-language model calls are costlier than text.
As an illustrative datapoint, the Manus agent reportedly costs on the order of \$2 USD in API usage to complete a single complex task on average.
This cost may be unacceptable at scale or for consumer use-cases.

\item \textbf{Challenge-4: Execution Performance.}
In many interactive scenarios, today's agents are significantly slower than human operators.
Complex tasks that require an agent to reason or plan over multiple steps are often bottlenecked by the LLM's inference speed and the overhead of iterative prompt exchanges\cite{sui2025, liu2025efficientinferencelargereasoning}.

\end{enumerate}

\feh{Existing model-based solutions still exhibit significant limitations. 
Regarding model safety, methods such as RLHF\cite{safe-rlhf, tan2025} are commonly employed to align the model securely. 
However, recent studies\cite{west2025} have demonstrated that these safety mechanisms can be circumvented merely by using specific prompts. 
Other approaches attempt to enable the model to self-correct\cite{zhao2025absolutezero, wang2025reinforcementlearningreasoninglarge}, but this process lacks any formal safety guarantees. 
On the other hand, to enhance the execution efficiency of agents as well as minimize the model size, 
current state-of-the-art methods often rely on model distillation\cite{zhang2025, dey2023, gemma2025, gu2024minillm} or pruning\cite{sun2024, ling2024, lee2025, hou2025pruning} to design specialized lightweight models for different agents. 
Nevertheless, such approaches constrain the applicability of agents to specific scenarios; 
when tasks or requirements change, the distilled or pruned models are unable to adapt.
Therefore, relying solely on model itself makes it challenging to simultaneously address the requirements for 
high reliability, privacy protection, performance efficiency, and cost-effectiveness in agent scenarios.}

In this paper, we propose \textbf{\sys (Agent Record \& Replay)}, a new paradigm inspired by the \emph{record-and-replay (R\&R)} techniques long used in software systems for debugging and reliability.

\begin{figure*}[t]
    \centering
    \includegraphics[width=0.98\linewidth]{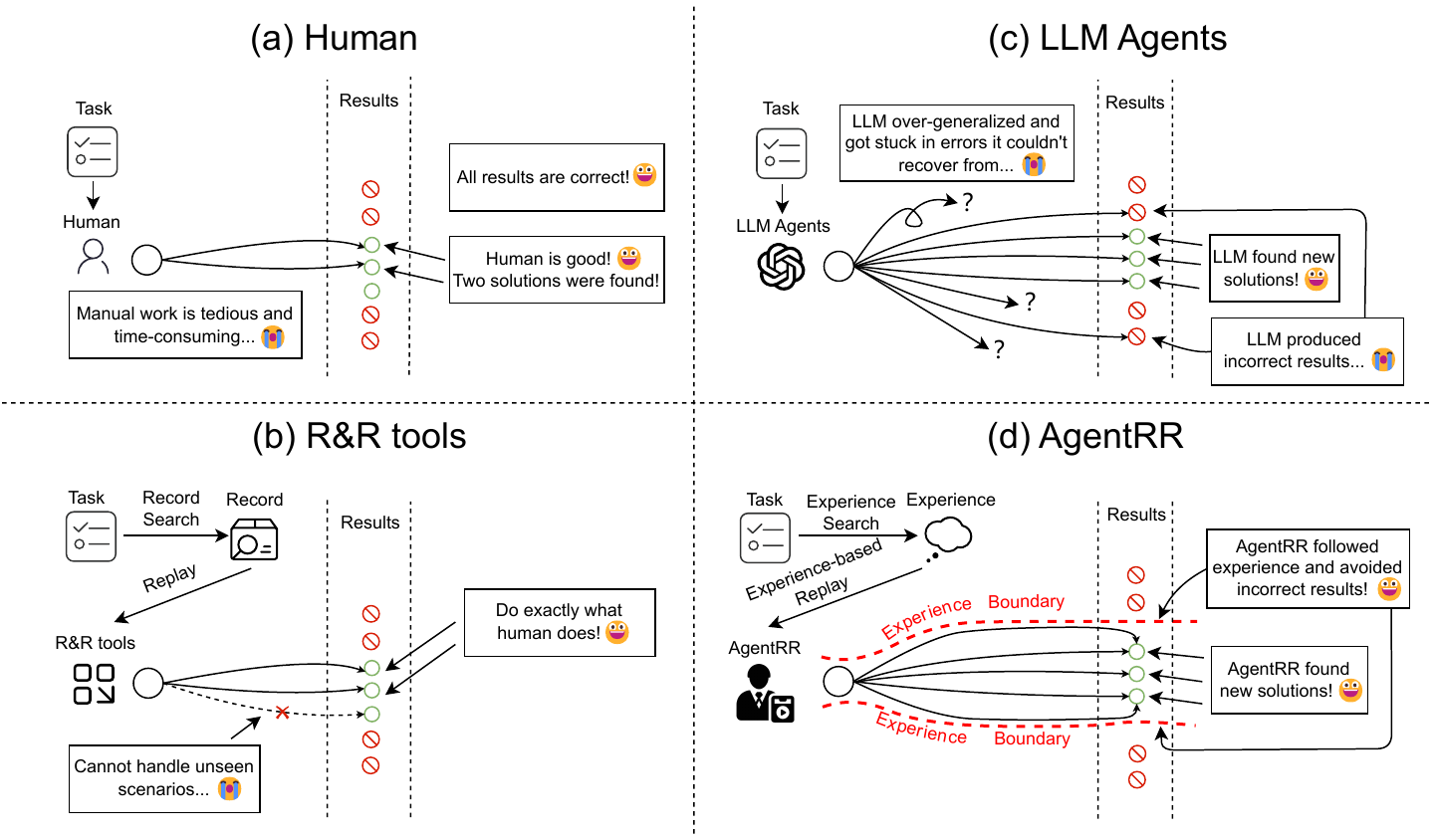}
    \caption{Conceptual comparison of human, R\&R tools, LLM agents, and AgentRR in task execution.}
    \label{fig:architecture}
\end{figure*}

Figure~\ref{fig:architecture} illustrates our envisioned system concept for AgentRR, contrasting it with human, traditional record-and-replay (R\&R) tools, 
and LLM agent approaches in handling tasks. 
In scenarios where humans can solve a task, they typically generate valid solutions and avoid incorrect results. 
R\&R tools, by searching a record database for human-generated traces and replaying them, 
can reliably reproduce these correct solutions. 
LLM agents, thanks to their strong generalization ability,
can not only accomplish tasks previously demonstrated by humans, 
but also derive novel solutions when tasks and environments change.
However, this generalization introduces two major risks: (1) LLM outputs are inherently probabilistic, 
so there is always a chance of producing incorrect results; 
(2) even with some self-correction capability, 
LLM agents may still enter unrecoverable error states 

AgentRR is guided by the design philosophy of \emph{bounded intelligence}: constraining agent intelligence within safe, successful experiences. By combining LLM capabilities with record-and-replay mechanisms, AgentRR leverages model generalization while grounding actions in proven experiences. This approach prevents unrecoverable errors and reduces incorrect outputs, while maintaining flexibility for valid new solutions.

After exploring the design philosophy above, we identify that the core to realizing this framework is the notion of \textbf{experience}. The experience abstraction is intended to \emph{decouple an agent's intelligence from its execution}: rather than requiring the agent to reason from scratch for every task, the agent can leverage accumulated knowledge from past successful behaviors. However, if an experience is too abstract, it offers strong generalization but may cause the replaying agent—especially if not sufficiently intelligent—to instantiate incorrect execution details or results, leading to replay failures. Conversely, if an experience is too concrete, it restricts the agent's generalization ability, making the experience difficult to reuse when the application scenario changes even slightly, which can also result in replay failure. Overall, naively using a single plan, trace, or prior LLM/human reasoning process as the experience is insufficient to achieve both reliability and generalization in practical agent systems.

To address these limitations, \sys introduces the concept of \textbf{multi-level experience}, which abstracts and summarizes past behaviors at different levels of granularity. \textbf{Low-level experiences} capture precise, concrete action sequences—such as specific UI operations or API calls—enabling rapid and reliable replay in environments highly similar to the original context. In contrast, \textbf{high-level experiences} provide more generalized procedural knowledge, allowing the agent to adapt to new environments or task variations by leveraging the LLM's reasoning ability to instantiate concrete actions from abstract plans. During execution, \sys dynamically selects the most appropriate level of experience based on the current environment and task requirements: when the context closely matches the original, low-level experiences are preferred for efficiency and reliability; when the context changes, high-level experiences offer greater flexibility and generalization. This multi-level experience framework allows \sys to balance the trade-off between reliability and generalization, ensuring that agents can both robustly reproduce known solutions and safely extend to new scenarios, all while maintaining strong safety guarantees through check functions at each level.

With the notion of multi-level experience, \sys adopts a workflow consisting of three main phases: \textbf{record}, \textbf{summary}, and \textbf{replay}. In the record phase, the system captures detailed traces of agent or human interactions while completing a task. The summary phase abstracts and organizes these traces into multi-level experiences, distilling both concrete action sequences and higher-level procedural knowledge. Importantly, during the summary phase, \sys also generates corresponding \textbf{check functions} for each experience. These check functions serve as safety boundaries, verifying execution flow integrity, state preconditions, parameter constraints, and other invariants to ensure that the agent's actions during replay remain reliable and secure. Finally, in the replay phase, the agent leverages the most suitable experience to guide its actions in new tasks, dynamically adapting to the current environment while ensuring safety and reliability through experience boundaries and check functions.

By structuring agent execution around multi-level experience-based R\&R, \sys fundamentally transforms the agent's operating mode. This paradigm shift brings significant benefits in reliability, privacy, cost, and performance, as detailed below:

\begin{itemize}

\item \textbf{Reduced reliance on LLM computation:}
Since most task steps can be executed by replaying a pre-derived plan (experience), the number of calls or the capability required by the LLM is drastically decreased.
This not only lowers the runtime cost of LLM invocation, but also reduces the power consumption of the agent which is especially beneficial for battery-powered devices.

\item \textbf{Improved reliability via guardrails:}
The structured experience with check functions contains the fixed workflow (sequence of actions) and constraints that were validated in a prior successful run.
Replaying this experience provides the agent with clear \emph{guardrails},
making its execution more deterministic and bounded. 
Errors stemming from LLM hallucinations can be prevented or caught early, 
because the agent is no longer free to arbitrarily diverge from the recorded template.

\item \textbf{Higher execution efficiency:}
Executing a predefined sequence from experience is typically much faster than dynamic reasoning.
Instead of spending time formulating each step via prompting and waiting for large-scale LLM responses, 
the agent can perform the next recorded action immediately or only needs to rely on a small-scale LLM for task replay.
The agent's performance becomes closer to that of a scripted program or human who already knows the procedure, rather than a model thinking from scratch.

\item \textbf{Enhanced privacy:}
When experiences are replayed mostly on the local device, and only minimal data need to be sent to the local LLM for interpretation, 
the exposure of user data is vastly reduced.
In \sys, one can imagine the heavy planning phase (record) being done in a secure environment or with a trusted model, and the replay phase running locally without transmitting raw user data.
Thus, \sys shifts the agent's operating mode in a way that can inherently strengthen privacy.
\end{itemize}

\sys is not a monolithic technique but rather a framework that can be realized in different ways to suit various scenarios.
The paradigm opens up a design space of flexible application modes, characterized by \emph{who (or what) performs the recording} and \emph{who performs the replay}.
By mixing and matching the entities responsible for record vs.\ replay, we can decouple ``intelligence'' from ``execution'' along different dimensions and optimize the use of resources and expertise:

\begin{itemize}
\item \textbf{User Record, Model Replay:}
In this mode, a human user demonstrates the task to be done (when the system is put in a `recording'' state).
The system captures the user's sequence of actions and outcomes in detail, then summarizes it into an experience representation that an agent can interpret and execute.
It is akin to \emph{programming by demonstration}: the user provides one successful example, and the agent can repeat that process later on new instances of the task.

\item \textbf{Large Model Record, Small Model Replay:}
Here, a powerful and expensive model is used to perform one or more representative runs of a task in a controlled setting, leveraging the large model's superior planning, understanding, and complex interaction capabilities to produce a correct solution trace.
Then, those experiences are deployed to resource-constrained edge devices (like smartphones or IoT devices) where a lightweight local agent can replay.
This enables low-cost, fast, and offline-capable execution, assuming the new task instance is sufficiently similar to the recorded experience.

\item \textbf{Untrusted Model Record, Trusted Model Replay:}
In this mode, one might use an untrusted model (e.g., on an untrusted cloud) to explore and generate a candidate solution for a task in a sandbox environment.
That exploratory run is recorded as an experience, potentially along with checkpoints to verify correctness.
Then, a small agent in a trusted environment (e.g., in TEE) replays the experience in the real environment, ensuring that only the vetted actions occur.
This effectively separates the model's ``exploration and discovery'' ability from the ``trusted execution'' process, allowing careful use of advanced model capabilities without compromising on critical safety or privacy policies in deployment.

\end{itemize}

Table~\ref{tab:rr_paradigms_experience} shows more possible scenarios of applying the Record \& Replay paradigm to LLM agents by outlining different combinations for who or what performs the recording and replaying actions,
along with their primary use cases.
When one LLM records for the same LLM to replay, it can lead to deterministic execution and skill consolidation.
If one LLM records for another LLM, this facilitates knowledge transfer and multi-agent coordination.
A small LLM recording for a large LLM can be used for trace verification and experience refinement.
Finally, a trusted LLM can record for an untrusted LLM to enable the safe exploration of new solutions.

\begin{table*}[htbp]
    \centering
    \caption{Comparison of Different R\&R Paradigms for LLM Agents}
    \label{tab:rr_paradigms_experience}
    \footnotesize 
    \begin{tabularx}{\textwidth}{@{} p{3cm} p{3cm} X @{}}
    \toprule
    \textbf{Recorder} & \textbf{Replayer} & \textbf{Primary Use Cases} \\
    \midrule
    Human & LLM & Task automation, knowledge capture from human experience\\
    \addlinespace
    LLM Agent & Human & Understanding agent behavior, teaching   \\
    \addlinespace
    One LLM & Same LLM & Deterministic execution, skill consolidation via experience replay  \\
    \addlinespace
    One LLM & Another LLM & Knowledge transfer via shared experience, multi-agent coordination   \\
    \addlinespace
    Large LLM & Small LLM & Bridging capability gaps, execution efficiency\\
    \addlinespace
    Small LLM & Large LLM & Trace verification, experience refinement \\
    \addlinespace
    Untrusted LLM & Trusted LLM & Privacy preserved execution\\
    \addlinespace
    Trusted LLM & Untrusted LLM & Safe exploration of new solutions\\
    \bottomrule
    \end{tabularx}
\end{table*}

Beyond these modes, \sys also paves the way for an ecosystem of experience sharing.
Once experiences are recorded and summarized, they could be standardized and packaged into modules that can be exchanged among users, devices, or applications.
We envision an \textbf{Experience Store} or \textbf{skill marketplace} where users can download or subscribe to experiences created by others, or contribute their own.
Shared experiences, especially if accompanied by metadata about their conditions of validity, could significantly reduce the cost and expertise required for individuals to deploy effective agents.

While \sys offers a promising direction to address many challenges of current agents, it is \emph{not a panacea}.
In tackling the old problems, it introduces new research questions that reflect the transformation of dynamic AI behavior into reusable knowledge artifacts.
Key among these are: How to ensure the \textbf{completeness} of recorded information so that the experience captures everything needed to reliably replay later?
How to maintain \textbf{robustness} of the replay in the face of environment changes or unexpected divergences?
How to improve the \textbf{generalizability} of experiences so that they can apply to a broader range of scenarios beyond the exact recorded case?
And how to delineate the \textbf{scope} of tasks for which R\&R is most suitable versus where a more flexible approach is needed?
The remainder of this paper delves into the \sys paradigm in detail, describes its implementation mechanisms, compares it to related paradigms, and discusses these open challenges and future directions.

\section{Related Work}

\subsection{Agent Reliability and Privacy}

The rapid advancement of AI agents has brought significant challenges in reliability and privacy. 
In terms of reliability, LLM agents often struggle with hallucinations --- generating plausible yet factually incorrect or nonsensical outputs. 
This tendency erodes user trust and can lead to substantial errors, particularly in scenarios involving chained actions where inaccuracies can accumulate and compound over time.
Moreover, the open-ended nature of LLMs makes their hallucination challenges distinct from earlier task-specific models, 
with factual errors being the predominant manifestation\cite{li2025survey, goldie2025}.

To address reliability issues, researchers have developed various approaches. 
Approaches such as LLM guardrails\cite{kang2024, han2025, deng2025, dong2024} aim to control an agent's inputs and outputs through multiple methods. 
RLHF\cite{safe-rlhf, tan2025}, for example, uses human judgments to shape a reward function for fine-tuning, 
which can reduce toxic or obviously incorrect outputs, but remains limited in preventing hallucinations or subtle errors. 
Retrieval-Augmented Generation (RAG) approaches\cite{gan2025retrievalaugmentedgenerationevaluation, li2025enhancingretrievalaugmentedgenerationstudy, singh2025agenticretrievalaugmentedgenerationsurvey} integrate real-time knowledge retrieval to ground responses in verified information, 
though their effectiveness depends heavily on the quality of retrieved content. 
Advanced prompting techniques like Chain-of-Thought (CoT)\cite{sui2025, chen2025} and ReAct prompting\cite{yao2023reactsynergizingreasoningacting} help structure the reasoning process and improve accuracy. 
Fact-checking mechanisms\cite{galichin2025icoveredbaseshere} help verify the accuracy of generated content. 
Additionally, in multi-agent systems\cite{chang2025sagallmcontextmanagementvalidation, zhou2025sweetrltrainingmultiturnllm}, 
employing multiple agents to check each other's work or vote on outcomes can improve reliability. 
However, this approach often incurs greater computational cost and system complexity.

Privacy concerns\cite{kim2025llmsonlineemergingthreat, dewitt2025openchallengesmultiagentsecurity, pinachodavidson2025proposalevaluatingoperationalrisk} 
arise particularly in cloud-hosted LLM agents, where uploading user data introduces risks of unauthorized access and potential misuse for personal secrets. 
The centralized nature of data handling in cloud environments, coupled with users' reduced control over their own information, 
underscores the urgency of addressing these privacy vulnerabilities. 
Research\cite{chen2025obviousinvisiblethreatllmpowered, chen2025clearcontextualllmempoweredprivacy} has demonstrated that LLMs can inadvertently memorize and subsequently reproduce sensitive information present in their training data, 
posing an additional privacy threat. 

To address these privacy challenges, researchers have developed various privacy-preserving techniques. 
Differential privacy\cite{charles2024finetuninglargelanguagemodels, li2022largelanguagemodelsstrong, wu2025} provides protection by adding statistical noise to individual data points. 
Federated learning\cite{mao2025privacypreservingfederatedembeddinglearning, ni2025federatedintelligencelargeai} enables collaborative training without centralizing sensitive data, 
while on-device processing\cite{zhang2025} eliminates the need for cloud transmission entirely. 
In addition, secure gateways\cite{li2025acesecurityarchitecturellmintegrated, bagdasarian2024} serve as intermediaries to detect and filter sensitive information, 
and machine unlearning techniques\cite{geng2025comprehensivesurveymachineunlearning} facilitate the removal of specific data points from trained models. 
Beyond the aforementioned algorithmic and framework-based approaches to privacy protection, 
a more direct solution involves running models locally, 
thereby preventing potential data breaches from malicious cloud service providers who might attempt to access users' private information.

\sys is compatible with all the aforementioned mechanisms. 
Moreover, through its experience-based record and replay approach, 
it significantly reduces the computational overhead of large language models,
enabling the replay phase for specific tasks to be executed entirely on resource-constrained edge devices under user control.

\subsection{Agent Performance and Cost}

The performance and operational cost of AI agents have emerged as significant challenges in their widespread adoption. 
Current agent systems inherit several limitations from their foundational LLMs, 
including limited context windows, difficulties in long-term planning and adaptation, inconsistent outputs, and susceptibility to hallucinations\cite{Huang2025, banerjee2024, CometLLMHallucination2025, xu2025}. 
These limitations often lead to substantial computational overhead and financial costs, 
particularly in complex tasks requiring multiple rounds of LLM interactions. 
Multi-modal tasks, such as GUI understanding and manipulation, are especially expensive due to the higher cost of vision-language model calls\cite{OpenAIPlatformPricing2025, GoogleGeminiPricing2025}.

Recent work has attempted to address these challenges through various approaches. 
Model distillation\cite{zhang2025, dey2023, gemma2025, gu2024minillm} and pruning methods\cite{sun2024, ling2024, lee2025, hou2025pruning} aim to reduce model size while maintaining performance. 
Parameter-efficient fine-tuning (PEFT) methods\cite{hu2021loralowrankadaptationlarge, wang2025tinatinyreasoningmodels, kong2024loraswitchboostingefficiencydynamic, sheng2024sloraservingthousandsconcurrent} enable task-specific adaptation with minimal parameter updates. 
Model quantization techniques\cite{wang2025bitnetv2native4bit, zheng2025empiricalstudyqwen3quantization, duanmu2025mxmoemixedprecisionquantizationmoe} reduce precision and memory footprint, 
potentially lowering computational costs. 
However, these approaches often trade off model capabilities for efficiency, potentially limiting the agent's ability to handle complex or novel tasks.

Other solutions focus on optimizing the interaction patterns between agents and LLMs. 
Task decomposition strategies\cite{sun2025fastslowthinkingcomplextasksolving, ren2025deepseekproverv2advancingformalmathematical, goldie2025} break down complex tasks into smaller, 
more manageable components, allowing the use of less expensive models for simpler subtasks. 
Caching mechanisms\cite{lin2025, rodionov2025hogwildinferenceparallelllm, chhikara2025} store and reuse previous LLM outputs to avoid redundant computations. 
Multi-agent architectures\cite{shao2025divisionofthoughtsharnessinghybridlanguage, wang2025dartllmdependencyawaremultirobottask} employ cost-efficient designs where cheaper LLMs handle initial processing while more expensive models are reserved for critical reasoning steps. 
However, these solutions still rely heavily on expensive model calls for each step of operation, and the fundamental challenge remains: 
how to maintain high performance while reducing the dependency on costly LLM computations.

In contrast, \sys offers a unique approach to address both performance and cost challenges through its record-and-replay mechanism. 
By capturing and reusing high-quality experiences from the most capable agents or human demonstrations,
\sys enables all agents to achieve performance comparable to the best performers without requiring expensive model calls for each execution. 


\subsection{Record \& Replay (R\&R) in Systems}

Record \& Replay techniques capture the execution of a program and allow its faithful re-execution later.
The core insight is that most program execution is deterministic;
only non-deterministic events (e.g., external inputs, thread scheduling, system call results, signals) need to be logged during recording.
During replay, the program is re-executed, and the logged non-deterministic inputs are injected at the appropriate points to reproduce the original execution state precisely.

R\&R is widely used for debugging hard-to-reproduce bugs\cite{song2025mimicme,Lam2017} (heisenbugs), security forensics\cite{Shalabi2018} (e.g., Backtracker), fault tolerance through replication,
and even enhancing hardware security mechanisms (e.g., RnR-Safe\cite{tournoux2011density}).
Tools like the rr debugger\cite{RRDebugger2025} achieve this for user-space applications
using techniques like ptrace for capturing syscalls/signals, seccomp-bpf to reduce interception overhead,
hardware performance counters for timing, and single-threading execution to avoid data races.
Other approaches involve VM-level recording\cite{deOliveira2006}, kernel modules\cite{Parasyris2023}, or application-integrated recording for distributed systems\cite{EECS20244} (e.g., aiRR)
which may selectively record events to reduce overhead.

In the field of Robotic Process Automation (RPA), Record \& Replay (R\&R) is also crucial for user interaction and browser automation. An RPA bot essentially records a user's sequence of actions—like data entry or form filling—and faithfully replays them to complete the task automatically. For example, tools like Playwright\cite{playwright2025} excel at recording and replaying user actions (e.g., clicks, text input, navigation), proving invaluable for automated testing to create reliable test suites mimicking real user journeys. This landscape is further evolving with the integration of AI agents into RPA. The browser-use system\cite{browser_use2024} aims to make websites accessible for AI agents by enabling workflows to be recorded and replayed, even with changes to web pages. WeReplay\cite{feng2023} employs deep learning models to infer rendering completion and optimally schedule the next replay event, demonstrating significant improvements in both replay accuracy and efficiency across devices. MobileGPT\cite{MobileGPT2024} also explores record and replay for LLM agents, using a hierarchical memory to store modular sub-tasks and replaying them when similar instructions are encountered, with adaptation to new contexts via pattern matching and few-shot learning. Systems like UFO2\cite{UFO22025} further advance this paradigm by introducing a comprehensive AgentOS architecture that integrates record and replay capabilities with deep OS-level integration. 
Recently, Workflow Use\cite{workflow-use} also applied record \& replay to make Browser Use more reliable and deterministic.
While this aligns with the direction of \sys, a notable distinction is that Workflow Use focuses on generating scripts to replay identical workflows, rather than summarizing general, multi-level experience.

\sys leverages the mechanism of recording user actions and replaying them through the agent,
but its goal differs significantly from traditional R\&R.
Traditional R\&R prioritizes fidelity\textemdash exact reproduction of a specific past execution instance.
The record phase in \sys captures user actions, but the aim is not bit-perfect replay.
Instead, the summary phase introduces generalization,
creating an abstract ``Experienc'' that represents a class of safe executions.
The replay phase enforces conformance to this generalized experience, not replication of a single trace.
The non-determinism experience-based R\&R must handle during replay is primarily the agent's creative output within designated nodes,
which is intentionally not constrained by the replay mechanism itself.

Table \ref{tab:comparison} summarizes the key differences between pure LLM agents, traditional R\&R tools, and \sys in terms of execution efficiency, accuracy, and generalization capabilities. 
Pure LLM agents excel in generalization but often suffer from low execution efficiency and inconsistent accuracy due to their reliance on expensive model calls and potential hallucinations. 
Traditional R\&R tools like Playwright achieve high accuracy and efficiency through deterministic replay, 
but their generalization capabilities are limited by their strict adherence to recorded actions. 
\sys combines the strengths of both approaches: it maintains high accuracy through experience-based replay, 
achieves efficient execution by reducing expensive model calls, 
and enables generalization through its structured experience management and summary mechanisms.

\definecolor{darkgreen}{rgb}{0.0, 0.5, 0.0} 

\begin{table}[t]
\centering
\caption{Comparison of different approaches in terms of execution efficiency, accuracy, and generalization capabilities.}
\label{tab:comparison}
\begin{tabular}{lccc}
\toprule 
\textbf{Approach} & \textbf{Execution Efficiency} & \textbf{Accuracy} & \textbf{Generalization} \\
\midrule 
Pure LLM Agent & \textcolor{red}{Low} & \textcolor{red}{Low} & \textcolor{darkgreen}{High} \\ 
Traditional R\&R & \textcolor{darkgreen}{High} & \textcolor{darkgreen}{High} & \textcolor{red}{Low} \\ 
\sys & \textcolor{darkgreen}{High (Exceeds human speed)} & \textcolor{darkgreen}{High} & \textcolor{darkgreen}{High (Generalized for repetitive tasks)} \\ 
\bottomrule 
\end{tabular}
\end{table}

\section{AgentRR: Multi-level Experience-driven Agent with Record \& Replay}


\feh{Most of contemporary agent systems widely adopt powerful cloud-based LLMs, 
such as GPT-4o, Claude 3.7, and Gemini 2.5 Pro. 
This reliance is primarily to facilitate complex task planning and GUI understanding. 
However, these LLM-based agent systems usually encounter inherent limitations, 
including high execution costs, unreliable execution outcomes, and potential privacy leakage concerns.}

\feh{To address the aforementioned challenges in current agent systems, 
this paper proposes \sys, a novel Record \& Replay mechanism specifically designed for agent applications. 
The design philosophy of \sys is inspired by human learning patterns, 
where individuals typically improve their accuracy and efficiency in performing similar tasks 
after observing demonstrations by others or through personal experience. 
Following this principle, AI agents should adopt a similar approach by recording historical operations and summarizing them into experiences 
to facilitate subsequent task execution. 
Compared to AI agents that solely rely on LLMs, 
\sys leverages system-level Record \& Replay mechanisms, achieving significant improvements in performance, reliability, and security.}


\subsection{Challenges in the Existing Record and Replay System}

\feh{However, unlike the traditional record-and-replay process, 
in agent-based scenarios, the environment and task requirements often exhibit the uncertainty and variability.
Consequently, the environment and tasks encountered during the replay phase typically differ from those in the recording phase. 
Applying precise record-and-replay methods tends to lead to a lower success rate and generalization capability in agent execution.}

\feh{To address this issue, a feasible approach is to abstract and summarize past behaviors into more generalized experiences.
In some prior work\cite{chhikara2025, shi2025, zhao2025,huang2025r2d2}, such experiences are often represented as concrete plans that the agent system can reference during execution. 
Although incorporating such experiences enhances the agent's generalization capabilities, 
it also introduces challenges related to the reliability of agent behavior due to the inherent uncertainty of generalized policies.
Therefore, balancing the trade-off between generalization and reliability in the agent's record-and-replay process is a critical problem that needs careful consideration.}

\subsection{Experience-centric Agent Execution}
To balance the reliability and generalization capabilities during the agent's execution,
we propose an experience-centric agent execution framework that encompasses two core components: 
multi-level experiences and check functions.

\subsubsection{Multi-level Experiences}
\begin{figure*}[htp]
    \centering
    \includegraphics[width=\linewidth]{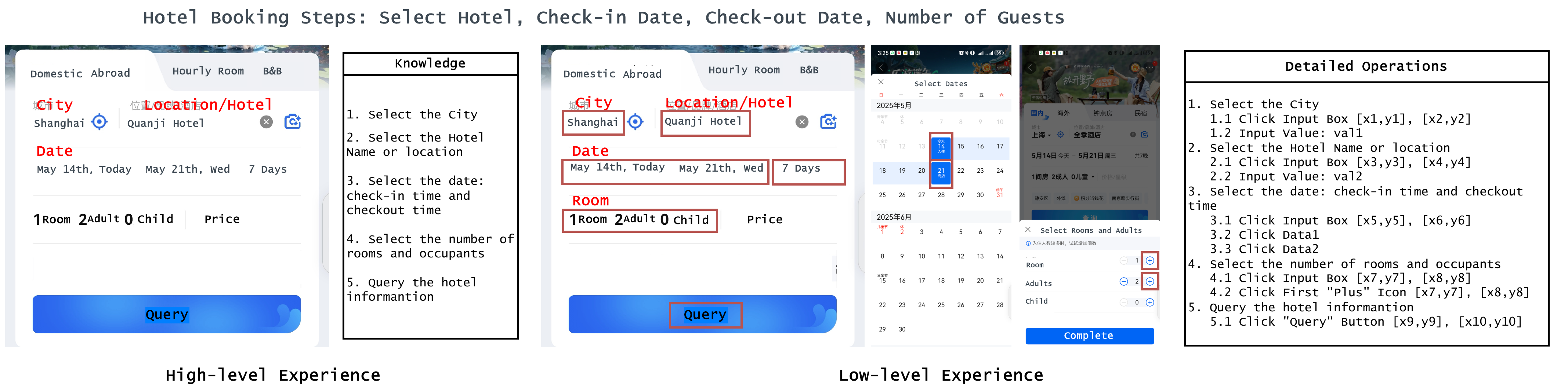}
    \caption{\textbf{Multi-level experience: }High-level experience describes the task planning process without being bound to specific platforms or UI layouts. 
    Low-level experience contains more detailed action decomposition and may be coupled with specific platforms and UI layouts.}
    \label{fig:design-experience}
\end{figure*}

\feh{While several prior agent systems~\cite{UFO22025, kwon2025} have proposed experience abstraction to enhance task completion, 
these approaches primarily focus on improving task success rates without thoroughly investigating 
how experiences can enhance agent execution efficiency and ensure operational reliability.
In our work, we introduce the novel concept of multi-level experiences.
The multi-level experience provides summarized knowledge with varying degrees of generalization, as shown in Figure~\ref{fig:design-experience}.
\textbf{Low-level experiences} offer more precise behavioral descriptions, 
enabling the agent system to replay these experiences more rapidly. 
However, these low-level experiences demonstrate limited generalization capability, 
as they are only effective in environments and tasks with high similarity to the original context.
For instance, they require the UI layout of applications to remain unchanged.
In contrast, \textbf{high-level experiences} represent more generalized summaries of historical records, 
typically without constraints on specific environments or exact procedural steps. 
During the replay phase, the high-level experience always requires a local model, 
and integrate the current environment and task context to generate concrete execution actions. 
Compared to executing from scratch, the high-level experience provides more prior knowledge, 
thereby reducing the reliance on the model's capacity in the replay stage.
\sys selects the appropriate level of experience for replay based on the current execution environment and task requirements.
For instance, when the agent's execution platform and applications remain identical to those in which the experience was generated,
\sys tends to utilize low-level experiences for replay operations.
However, when the platform or applications undergo changes, to ensure sufficient generalization capability,
\sys opts for high-level experiences to facilitate better planning during the agent's execution process.
}

\subsubsection{Check Functions}
\feh{On the other hand, since experiences are not concrete action trajectories, 
the system relies on local LLMs to translate experiences into specific action sequences for agent execution.
However, due to the inherent hallucination problems and insufficient training of LLMs,
the reliability of the agent system may significantly deteriorate and potentially violate user safety requirements.
To ensure reliability and security during the experience-centric replay phase,
we define distinct check functions for experiences at different levels. 
These check functions delineate the boundaries of the generalization capability of experiences and serve as a TCB (Trusted Computing Base) during the agent's execution.
Check functions can be generated in various ways; 
for instance, they may be explicitly defined by the user as specific code implementations. 
Or, they may be generated by the user-provided description and ML-based summary.
Once the user audits and places trust in the given check functions, 
it guarantees that the agent will adhere to the user's safety requirements throughout the replay process.}

More specifically, the check function typically verifies:

\begin{itemize}[leftmargin=*]
    \item \textbf{Execution Flow Integrity:} The integrity of the agent's execution process is formally defined 
    to verify that the agent does not transition into undefined states.
    For instance, in an order processing agent, making payments without user confirmation would be considered an illegal operation.
    For such cases, the check function can verify LLM's execution actions by obtaining application interface information and 
    leveraging OCR capabilities.
    Or, at the system level, it can prevent unauthorized payment requests by blocking related network transmissions.
    \item \textbf{State Preconditions:} Preconditions associated with the action should meet the requirement.
    For example, in a form-filling agent system, there are often dependencies between form fields. 
    Therefore, the check function needs to verify that all prerequisite fields meet the required conditions 
    before proceeding with subsequent field entries.
    \item \textbf{Data/Parameter Constraints:} 
    If the action involves parameters, 
    these parameters should consistent with constraints specified in the Experience.
    Since user tasks often involve different inputs, specific outputs are not predefined in the experience. 
    Therefore, the check function needs to verify whether the values generated during agent execution align with those defined in the user task.
    \item \textbf{Safety Invariants:} Check functions can enforce safety invariants identified during Summary or specified manually. 
    This addresses the requirement that variable parts of an Experience should not compromise safety. 
    For example, loop operations are common task patterns in agent scenarios. 
    While the number of iterations may not compromise the integrity of the Execution Flow, 
    it can significantly impact the correctness of agent task execution. 
    Therefore, check functions must implement dedicated verification mechanisms for loop operations.
\end{itemize}

\subsection{Experience Store}

The Experience Store acts as a central repository for managing different levels of Experiences.
It allows users to upload Experiences generated from their traces, download Experiences created by others, search for Experiences relevant to specific tasks, 
rate the quality and reliability of Experiences, and potentially mark Experiences as formally audited or verified.
The store is likely implemented as a database storing the Experience organized as JSON or a graph database format, 
along with associated metadata (task description, creator, version, ratings, audit status, usage statistics).
It provides an interface for users or agents to query and select the most appropriate Experience for a given task based on criteria like task similarity, user ratings, success rate, etc.

\subsection{State Transition Diagram}
\feh{To better describe the process of agent record and replay, 
we represent the agent's operations as a trajectory within a state transition diagram. 
In this diagram, nodes correspond to environment states, 
and edges correspond to actions executed by the agent. 
When an agent completes a task, it is equivalent to obtaining a complete trajectory within the state transition diagram.
More specifically:
}

\textbf{States (S):} The state represents a snapshot of the relevant aspects of the system environment at a given point in time. 
The definition of ``state'' is flexible and task-dependent. 
It could include: the currently focused application and window, 
the status or content of specific UI elements, relevant file system information, 
or even abstract states derived from these (e.g., "Logged In," "File Open"). 
Furthermore, states can be defined at varying levels of granularity. 
In the context of multi-level experiences, 
multiple low-level states can be aggregated into a single high-level state.
\feh{A well-defined state is fundamental in record and replay systems. 
It enables accurate comparison of the states of different nodes,
and can determine the feasibility of subsequent actions.
}

\textbf{Actions (A):} Actions correspond to the procedural operations represented by the edges in the State Transition Diagram, 
such as \texttt{click(button\_id)}, \texttt{type(text\_field\_id, 'text')}, \texttt{call\_api(endpoint, params)}.
\feh{Furthermore, the agent system can effectively minimize the complexity of state transitions within the state transition diagram by specifying a limited set of meta-operations.
}

\textbf{Transitions:} A transition $S \xrightarrow{A} S'$ denotes that executing action $A$ in state $S$ leads to state $S'$. 
The state transition diagram defines all allowed transitions.

\textbf{Trajectories:} A complete execution of a task during replay/record phase corresponds to a trajectory in the state transition diagram, 
which is a sequence of states and actions: $S_0 \xrightarrow{A_1} S_1 \xrightarrow{A_2} S_2 \cdots \xrightarrow{A_n} S_n$. 
The goal of \sys is to ensure that this trajectory is valid according to the chosen experience.

\feh{The model's role in this process is to \textbf{select the most appropriate path} from among all possible trajectories for execution. 
Since each environment state offers a large number of latent executable actions, 
this leads to high demands on the model's capability and results in longer inference latency.
Experience is defined as a summarized collection of trajectories from similar tasks, 
effectively representing a small set of trajectories. 
By leveraging experience, the search space for the model's action selection is significantly reduced, 
thereby improving the agent's efficiency. 
However, experience also constrains the agent's generalization ability, 
with its behavior boundaries regulated by a check function.
}

\section{Record\&Replay Modules in the \sys System}

\begin{figure*}[htp]
    \centering
    \includegraphics[width=\linewidth]{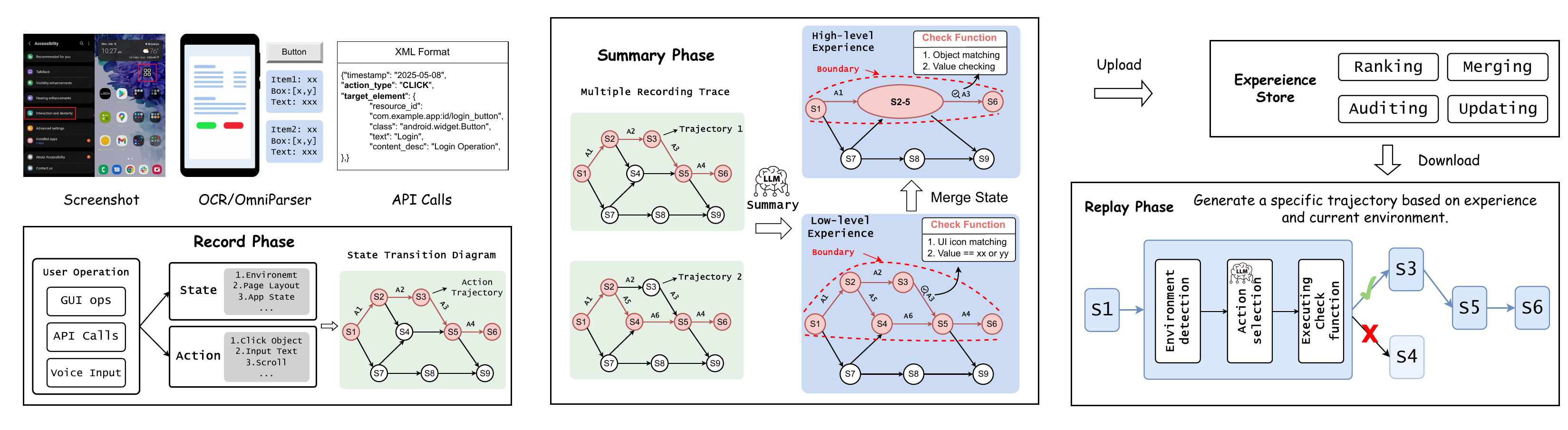}
    \caption{\textbf{The overall architecture of AgentRR: }The AgentRR system consists of three core components:
    the Record module, Summary module, and Replay module. 
    Additionally, to facilitate experience sharing across different users, 
    AgentRR incorporates an experience store.}
    \label{fig:design-overall}
\end{figure*}

Figure~\ref{fig:design-overall} illustrates the overall architecture of AgentRR.
The record phase is responsible for capturing user interactions and generating detailed action traces.
The summary phase is responsible for summarizing the action traces into reusable experiences.
The replay phase is responsible for guiding the agent's execution according to a selected experience.

\subsection{Record Phase: Capturing User Traces}
\feh{The Record phase requires capturing a complete sequence of actions, 
which can be accomplished by logging either GUI interactions or API calls. 
To ensure a high success rate during the subsequent replay phase, 
the information recorded during the record phase is critically important and primarily consists of two aspects:
}

\feh{First, the state of the environment at each step needs to be recorded. 
Since each state may contain a large amount of information, 
there is a necessary trade-off between the amount of data recorded and the successful rate during the replay phase. 
A practical recording strategy involves detailed recording of key elements that influence the current behavior, 
while recording non-critical elements more coarsely. 
For example, in GUI operations, the layout of the page can be recorded in a simplified manner, 
whereas the elements involved in the interactions should be recorded with as much detail as possible.
Second, the Record phase must capture the operations that cause state transitions. 
To accurately record these operations, 
a set of predefined meta-operations can be used, 
such as clicking a specific object, entering certain input, 
sliding by a certain proportion, or directly invoking an API call. 
Through careful design, a combination of these meta-operations can be used to accomplish a given task.
}

\feh{In summary, the record phase generates a detailed action trace by recording the environment state and user actions at every step, 
resulting in a path within the state transition diagram.
}

\subsection{Summary Phase: Generalizing Traces to Experience}

This phase transforms one or more raw execution traces into a generalized, reusable experience graph.
This is arguably the most challenging and critical phase for ensuring both utility and safety.
Traces for the same task may exhibit variations due to user choices, different starting states, or minor environmental changes.
\feh{Therefore, during the summary process, whether performed by a ML model or manually, 
it is essential to identify the commonalities into the experience,
but delegate the differences to be handled during the replay phase. 
Moreover, to address the abstraction of multi-level experiences, 
summarization can be conducted at varying granularities.
For instance, experiences can be categorized either at the level of precise operations 
or as high-level action plans.
}

\feh{In the context of a state transition diagram, 
the summary process can be viewed as templated actions from trajectories of similar tasks. 
Starting from a given node, these templated actions enable transitions to nodes which have similar states.
For multi-level experiences, a high-level experience is created by merging states from multiple low-level experiences,
enabling the high-level experience to adapt to more generalized application scenarios.
}

\feh{In addition to summarizing the commonalities among similar operations, 
the summary phase also needs to generate the corresponding check function. 
Similarly, the check function itself can be precisely defined by the user 
or generated by the model combined with the user's behavior. 
The check function constitutes the trusted computing base (TCB). 
We can ensure the trustworthiness of the check function through manual audits, 
bug bounty programs, and other verification methods.
The form of the check function is flexible; 
it can be a piece of precise verification code (e.g., disallowing clicks on security-sensitive elements), 
or a natural language description coupled with a validation model. 
Regardless which type of check function is used, 
it should be significantly smaller than the agent model size. 
This design helps reduce the size of the TCB within the overall agent system.
}

\subsection{Replay Phase: Agent Execution and Self Optimization}

\feh{During the replay phase, the agent system replays existing experience based on the current task and environment. 
Since the replay phase does not utilize exact action trajectory but rather experience generated through summarization, 
the agent must possess two core capabilities during execution.
First, the local model employed by the agent should be able to generate a series of concrete actions from the summarized experience, 
tailored to the current environment and task, 
and successfully execute it within the given environment.
Second, the agent must be capable of performing a check function and, 
based on its results, promptly prevent any unsafe operations.}

\feh{Beyond simply replaying previously accumulated experiences, 
an agent system should be capable of selecting the most suitable experience for a given task. 
As multi-level experience may exist for the same task, 
and different users may exhibit distinct operational behaviors, 
the agent system can achieve improvements in two key aspects.
First, low-level experiences tend to yield better performance during the replay phase; 
however, changes in the environment or task context can lead to failures when relying on these low-level experiences. 
Consequently, the agent system aims to select the lowest-level experience that still maintains the highest success rate during replay phase. 
Furthermore, for frequently executed tasks, 
the system incrementally refines its experience repository by continuously abstracting more fundamental or lower-level experiences. 
This process enables the agent to perform common tasks with increasing efficiency over time.
Second, when considering the varied experience from different users, 
the agent employs a ranking mechanism to identify and select the optimal experience for replay. 
The system can also incorporate high-quality experiences sourced from other users, 
merging them into its local experience repository to facilitate continuous improvement.}

\section{Case Study}

\begin{figure}[htbp]
\centering
\includegraphics[scale=0.15]{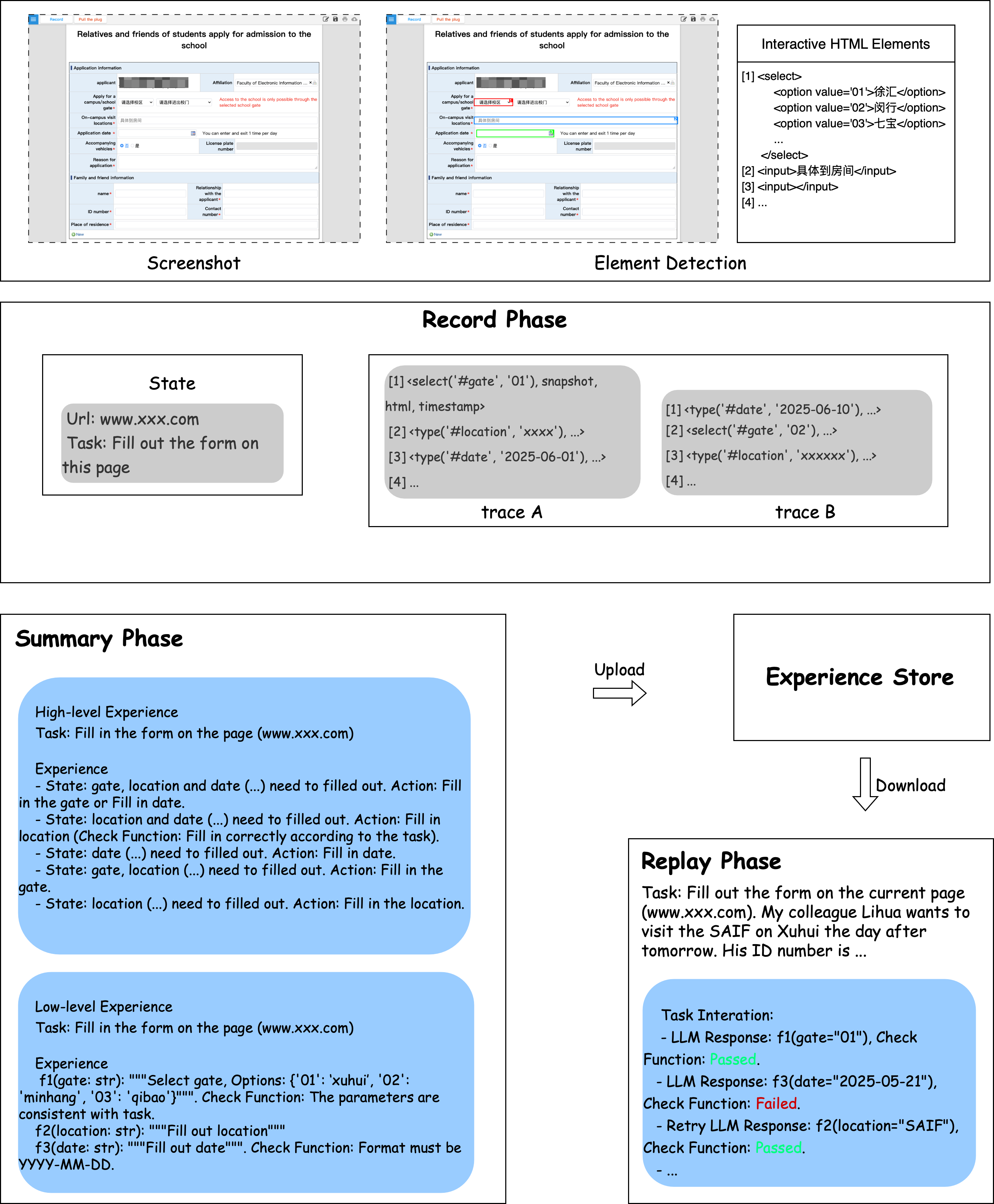}
\caption{The online form filling example: During the record phase, users capture multiple trace behaviors. 
In the summary phase, these traces are synthesized into multi-level experiences based on both scripts and natural language descriptions. 
During replay, the system selects the optimal experience and, 
in conjunction with the specific task requirements, controls web interactions to complete the form.}
\label{fig:case_2}
\end{figure}

To illustrate the practical effectiveness of AgentRR, 
we evaluate its performance on a real-world form-filling task (depicted in Figure~\ref{fig:case_2}), 
where existing agent-based solutions face significant challenges in terms of both accuracy and efficiency.
The SOTA LLM-based agent like OpenAI's CUA requires approximately three minutes to process this form, 
while still falling short of achieving the complete task.
Traditional record-replay tools (such as Chrome Recorder) lack generalization capabilities, requiring manual parameter input for each operation sequence.
In contrast, AgentRR enables users to simply describe the task in natural language, 
and LLM autonomously and quickly fills out the form based on the previously recorded experience.

\textbf{Record Phase }
As illustrated in the figure \ref{fig:case_2}, we present three form options in the table here for demonstration. 
During recording, users employ trace tools such as Chrome Recorder~\cite{ChromeRecorder2024} and Playwright Codegen~\cite{PlaywrightCodegen2024} to capture their interaction sequence, 
and then attach information such as the page URL, task descriptions and screenshot. 
In this example, the user records two traces, trace A fills the form fields in sequential order; trace B fills out of order.

\textbf{Summary Phase }
In the summary stage, the system first replays the recorded traces to verify their reproducibility. 
If errors occur — often due to dynamically generated HTML elements causing locator failures — the system collects diagnostic data.
Vaild traces and dignostic information are then processed by the LLM to summarize experience:

High-Level Experience is usually generated by LLMs, 
it describes the task's current state, next-step actions, and conditions for the check function to validate. 
Low-Level Experience typically consists of a series of API calls or executable scripts 
that usually can be directly derived from trace logs.
To achieve generalization capabilities, 
low-Level experience identifies variables within traces, 
along with their descriptions and associated constraints (e.g., check function).

\textbf{Replay Phase }
When users submit new tasks, AgentRR leverages the summarized experiences for execution.
For models with poor capabilities, Low-Level Experience fills partial content via parameterized APIs. 
When Low-Level Experience fails to meet task requirements, 
the system falls back to High-Level Experience for task execution, 
though this approach necessitates more powerful model capabilities.

The CheckFunction blocks executions that violate preconditions or deviate from recorded behaviors. 
For example, in the dependency checking of the check function,
the call \texttt{f3(date=xxx)} in figure \ref{fig:case_2} is technically feasible,
but is rejected since none of the valid previous traces demonstrated filling the ``date'' field after the ``gate'' field.
The system enforces strict ordering constraints in scenarios where arbitrary input sequences could potentially compromise the integrity of the task execution.

\section{Discussion}

Our design and evaluation demonstrate that \sys effectively addresses the challenges of agent reliability, privacy, and performance through its innovative record-and-replay mechanism. 
The multi-level experience design enables efficient task decomposition and long-term planning, while the check functions provide robust safety guarantees. 
However, several limitations and areas for future work warrant discussion.


\textbf{Summary Complexity:}
Summarizing recorded traces into multi-level experiences remains challenging for several reasons.
First, there is still a lack of clear definitions for different levels of experiences.
In our implementation, we define high-level experiences as knowledge representations, while low-level experiences correspond to UI operations or API calls.
For high-level experiences, specialized LLMs can be employed to generate meaningful summaries.
Low-level experiences can be directly transformed from recorded traces through dedicated conversion scripts.
However, these approaches still face limitations such as incomplete summarization and potential failures in generating executable scripts.

\textbf{State Space:}
The definition of states in the state transition diagram also varies across different tasks.
For UI-based agents, the states can incorporate UI elements and layouts of applications as integral components.
In contrast, for command-line and API-based agents, 
states can leverage existing operating system and filesystem states.
Additionally, adjacent states can be merged synchronously to encompass a broader range of scenarios.

\textbf{Generalization vs. Safety:}
This remains a fundamental tension.
The Summary phase must generalize to create reusable Experiences, 
but over-generalization can compromise the safety guarantees derived from the specific human traces.
Our current prototype relies on conservative generalization,
focusing primarily on repetitive tasks such as filling out web forms, making hotel reservations and conducting online purchases, etc.
These tasks are characterized by well-defined boundaries and exhibit similar execution flows across different requests.
In handling such tasks, \sys effectively balances agent safety with the ability to generalize across varying users' requests.


\textbf{Recording Fidelity and Robustness:}
Accurately recording user behaviors presents significant challenges.
In web-based scenarios, the presence of dynamic HTML elements and inconsistent HTML element usage by developers poses difficulties.
Even with tools like Browser-use and Playwright for web behavior recording, achieving 100\% reliable replay remains elusive.
In UI-based scenarios, challenges persist in accurately capturing user interface interactions due to dynamic pop-ups, varying screen resolutions, and other interface-related complexities.

\textbf{User Burden:}
Recording demonstrations requires initial user effort.
Furthermore, building trust in shared Experiences within the Experience Store likely requires mechanisms for auditing, rating, or verification, adding an additional layer of human oversight or computational cost.

\section{Conclusion}

This paper introduces \sys, a novel paradigm that addresses the fundamental challenges of modern AI agents through an experience-based record-and-replay (R\&R) approach. 
By decoupling an agent's intelligence from its execution, 
\sys offers practical solutions to the core challenges of reliability, privacy, operational cost, and execution performance. 
To achieve this, \sys proposes multi-level experience design, 
where lower-level experiences provide precise behavioral operations for rapid replay, 
while high-level experiences offer more generalized summaries for better adaptation to varying environments. 
To strengthen the reliability and safety of AI agents, \sys proposes check functions as a trusted computing base (TCB) 
that verifies execution flow integrity, state preconditions, and safety invariants, ensuring reliable and secure agent operations. 
Compared with other pure LLM-based agents, 
\sys opens up new research directions by integrating system-level record and replay capabilities, 
while leveraging experiences to achieve increasingly efficient and accurate agent execution.
\end{sloppypar}

\bibliographystyle{ACM-Reference-Format}
\bibliography{references}

\end{document}